# SOCKEYE: A Toolkit for Neural Machine Translation

**Felix Hieber, Tobias Domhan, Michael Denkowski,
David Vilar, Artem Sokolov, Ann Clifton, Matt Post**
{fhieber,domhant,mdenkows,dvilar,artemsok,acclift,mattpost}@amazon.com

## Abstract

We describe SOCKEYE,[1] an open-source sequence-to-sequence toolkit for Neural Machine Translation (NMT). SOCKEYE is a production-ready framework for training and applying models as well as an experimental platform for researchers. Written in Python and built on MXNET, the toolkit offers scalable training and inference for the three most prominent encoder-decoder architectures: attentional recurrent neural networks, self-attentional transformers, and fully convolutional networks. SOCKEYE also supports a wide range of optimizers, normalization and regularization techniques, and inference improvements from current NMT literature. Users can easily run standard training recipes, explore different model settings, and incorporate new ideas. In this paper, we highlight SOCKEYE's features and benchmark it against other NMT toolkits on two language arcs from the 2017 Conference on Machine Translation (WMT): English–German and Latvian–English. We report competitive BLEU scores across all three architectures, including an overall best score for SOCKEYE's transformer implementation. To facilitate further comparison, we release all system outputs and training scripts used in our experiments. The SOCKEYE toolkit is free software released under the Apache 2.0 license.

## 1 Introduction

The past two years have seen a deep learning revolution bring rapid and dramatic change to the field of machine translation. For users, new neural network-based models consistently deliver better quality translations than the previous generation of phrase-based systems. For researchers, Neural Machine Translation (NMT) provides an exciting new landscape where training pipelines are simplified and unified models can be trained directly from data. The promise of moving beyond the limitations of Statistical Machine Translation (SMT) has energized the community, leading recent work to focus almost exclusively on NMT and seemingly advance the state of the art every few months.

For all its success, NMT also presents a range of new challenges. While popular encoder-decoder models are attractively simple, recent literature and the results of shared evaluation tasks show that a significant amount of engineering is required to achieve "production-ready" performance in both translation quality and computational efficiency. In a trend that carries over from SMT, the strongest NMT systems benefit from subtle architecture modifications, hyper-parameter tuning, and empirically effective heuristics. Unlike SMT, there is no "de-facto" toolkit that attracts most of the community's attention and thus contains all the best ideas from recent literature.[2] Instead, the presence of many independent toolkits[3] brings diversity to the field, but also makes it difficult to compare architectural and algorithmic improvements that are each implemented in different toolkits.

---

[1] https://github.com/awslabs/sockeye (version 1.12)
[2] For SMT, this role was largely filled by MOSES [Koehn et al., 2007].
[3] https://github.com/jonsafari/nmt-list

To address these challenges, we introduce SOCKEYE, a neural sequence-to-sequence toolkit written in Python and built on Apache MXNET[4] [Chen et al., 2015]. To the best of our knowledge, SOCKEYE is the only toolkit that includes implementations of all three major neural translation architectures: attentional recurrent neural networks [Schwenk, 2012, Kalchbrenner and Blunsom, 2013, Sutskever et al., 2014, Bahdanau et al., 2014, Luong et al., 2015], self-attentional transformers [Vaswani et al., 2017], and fully convolutional networks [Gehring et al., 2017]. These implementations are supported by a wide and continually updated range of features reflecting the best ideas from recent literature. Users can easily train models based on the latest research, compare different architectures, and extend them by adding their own code. SOCKEYE is under active development that follows best practice for both research and production software, including clear coding and documentation guidelines, comprehensive automatic tests, and peer review for code contributions.

In the following sections, we provide an overview of SOCKEYE's three encoder-decoder architectures (§2). We then highlight key model, training, and inference features (§3) and benchmark SOCKEYE against a number of other open-source NMT toolkits on two public datasets, WMT 2017 English–German and Latvian–English [Bojar et al., 2017] (§4). Measured by BLEU score [Papineni et al., 2002], SOCKEYE's recurrent model is competitive with that of the best-performing toolkit, the convolutional model performs best in class, and the transformer model outperforms all other models from all toolkits. We conclude the paper with a summary and invitation for collaboration (§5).

## 2 Encoder-Decoder Models for NMT

Since the early days of neural sequence-to-sequence modeling, the encoder-decoder model has been appealing. The core idea is to *encode* a variable-length input sequence of tokens into a sequence of vector representations, and to then *decode* those representations into a sequence of output tokens. This decoding is conditioned on information from both the latent input vector encodings as well as its own continually updated internal state, motivating the idea that the model should be able to capture meanings and interactions beyond those at the word level.

Formally, given source sentence $X = x_1, ..., x_n$ and target sentence $Y = y_1, ..., y_m$, an NMT system models $p(Y|X)$ as a target language sequence model, conditioning the probability of the target word $y_t$ on the target history $Y_{1:t-1}$ and source sentence $X$. Each $x_i$ and $y_t$ are integer ids given by source and target vocabulary mappings, $\mathbf{V}_{src}$ and $\mathbf{V}_{trg}$, built from the training data tokens and represented as one-hot vectors $\mathbf{x}_i \in \{0,1\}^{|\mathbf{V}_{src}|}$ and $\mathbf{y}_t \in \{0,1\}^{|\mathbf{V}_{trg}|}$. These are embedded into $e$-dimensional vector representations, $\mathbf{E}_S \mathbf{x}_i$ and $\mathbf{E}_T \mathbf{y}_t$, using embedding matrices $\mathbb{R}^{e \times |\mathbf{V}_{src}|}$ and $\mathbf{E}_T \in \mathbb{R}^{e \times |\mathbf{V}_{trg}|}$. The target sequence is factorized as $p(Y|X;\boldsymbol{\theta}) = \prod_{t=1}^{m} p(y_t|Y_{1:t-1}, X; \boldsymbol{\theta})$. The model, parameterized by $\boldsymbol{\theta}$, consists of an *encoder* and a *decoder* part, which vary depending on the model architecture. $p(y_t|Y_{1:t-1}, X; \boldsymbol{\theta})$ is parameterized via a softmax output layer over some decoder representation $\bar{\mathbf{s}}_t$:

$$p(y_t|Y_{1:t-1}, X; \boldsymbol{\theta}) = \text{softmax}(\mathbf{W}_o \bar{\mathbf{s}}_t + \mathbf{b}_o), \qquad (1)$$

where $\mathbf{W}_o$ scales to the dimension of the target vocabulary $\mathbf{V}_{trg}$.

For parameterizing the encoder and the decoder, Recurrent Neural Networks (RNNs) constitute a natural way of encoding variable-length sequences by updating an internal state and returning a sequence of outputs, where the last output may summarize the encoded sequence [Sutskever et al., 2014]. However, with longer sequences, a fixed-size vector may not be able to store sufficient information about the sequence, and attention mechanisms help to reduce this burden by dynamically adjusting the representation of the encoded sequence given a state [Bahdanau et al., 2014]. While recurrent architectures with attention have become a de-facto standard for NMT in the last years, they exhibit the core limitation that the representation of a token in a sequence at position $i$ cannot be computed before all previous representations are known. With the rise of highly-parallel devices such as GPUs and the need for large amounts of training data for these neural architectures, parallelization of encoder/decoder computation has become paramount.

More recently, two additional architectures have been proposed to encode sequences, which have, besides of improving parallelization, improved the state of the art in NMT: the self-attentional transformer [Vaswani et al., 2017], and the fully convolutional model [Gehring et al., 2017]. We give a brief description for all three architectures as implemented in SOCKEYE, but refer the reader to the references for more details.

---

[4]https://mxnet.incubator.apache.org/

## 2.1 Stacked RNN with Attention

We start by defining the recurrent architecture as implemented in SOCKEYE, following Bahdanau et al. [2014] and Luong et al. [2015].

**Encoder** The first encoder layer consists of a bi-directional RNN followed by a stack of uni-directional RNNs. Specifically, the first layer produces a forward sequence of hidden states $\overrightarrow{\mathbf{h}}_1^0 \ldots \overrightarrow{\mathbf{h}}_n^0$ through an RNN, such that:

$$\overrightarrow{\mathbf{h}}_i^0 = f_{enc}(\mathbf{E}_S \mathbf{x}_i, \overrightarrow{\mathbf{h}}_{i-1}^0), \tag{2}$$

where $\overrightarrow{\mathbf{h}}_0^0 = \mathbf{0} \in \mathbb{R}^d$ and $f_{enc}$ is some non-linear function, such as a Gated Recurrent Unit (GRU) or Long Short Term Memory (LSTM) cell. The reverse RNN processes the source sentence from right to left: $\overleftarrow{\mathbf{h}}_i^0 = f_{enc}(\mathbf{E}_S \mathbf{x}_i, \overleftarrow{\mathbf{h}}_{i+1}^0)$, and the hidden states from both directions are concatenated: $\mathbf{h}_i^0 = [\overrightarrow{\mathbf{h}}_i^0; \overleftarrow{\mathbf{h}}_i^0]$, where the brackets denote vector concatenation. The hidden state $\mathbf{h}_i^0$ thus can incorporate information from both tokens to the left, as well as tokens to the right. The bi-directional RNN is follow by a stack of uni-directional layers. For encoder layer index $l > 1$, the input at position $i$ to $f_{enc}^l$ is the output of the lower layer: $\mathbf{h}_i^{l-1}$. With deeper networks, learning turns increasingly difficult [Hochreiter et al., 2001, Pascanu et al., 2012] and residual connections of the form $\mathbf{h}_i^l = \mathbf{h}_i^{l-1} + f_{enc}(\mathbf{h}_i^{l-1}, \mathbf{h}_{i-1}^l)$ become essential [He et al., 2016].

**Decoder** The decoder consists of an RNN to predict one target word at a time through a state vector $\mathbf{s}$:

$$\mathbf{s}_t = f_{dec}([\mathbf{E}_T \mathbf{y}_{t-1}; \bar{\mathbf{s}}_{t-1}], \mathbf{s}_{t-1}), \tag{3}$$

where $f_{dec}$ is a multi-layer RNN, $\mathbf{s}_{t-1}$ the previous state vector, and $\bar{\mathbf{s}}_{t-1}$ the source-dependent *attentional vector*. Providing the attentional vector as an input to the first decoder layer is also called *input feeding* [Luong et al., 2015]. The initial decoder hidden state is a non-linear transformation of the last encoder hidden state: $\mathbf{s}_0 = \tanh(\mathbf{W}_{init} \mathbf{h}_n + \mathbf{b}_{init})$. The attentional vector $\bar{\mathbf{s}}_t$ combines the decoder state with a *context vector* $\mathbf{c}_t$:

$$\bar{\mathbf{s}}_t = \tanh(\mathbf{W}_{\bar{s}}[\mathbf{s}_t; \mathbf{c}_t]), \tag{4}$$

where $\mathbf{c}_t$ is a weighted sum of encoder hidden states: $\mathbf{c}_t = \sum_{i=1}^n \alpha_{ti} \mathbf{h}_i$. The attention vector $\boldsymbol{\alpha}_t$ is computed by an attention network [Bahdanau et al., 2014, Luong et al., 2015]:

$$\alpha_{ti} = \text{softmax}(\text{score}(\mathbf{s}_t, \mathbf{h}_i))$$

where $\text{score}(\mathbf{s}, \mathbf{h})$ could be defined as (see Table 1 for more attention types):

$$\text{score}(\mathbf{s}, \mathbf{h}) = \mathbf{v}_a^\top \tanh(\mathbf{W}_u \mathbf{s} + \mathbf{W}_v \mathbf{h}). \tag{5}$$

Due to the recurrent dependency on the previous time step, the computation of RNN hidden states can not be parallelized over time. The dependencies in the computation graph lead to a triangular computation going from the bottom left to the top right, as each RNN cell needs to wait for both the previous time step of the same layer and the same time step of the previous layer (also see Figure 1a) On the decoder side, computation is even more serialized due to the *input feeding* mechanism, as the second time step can only be started once the first has been completed. Computation happens therefore one column at a time.

## 2.2 Self-attentional Transformer

The transformer model [Vaswani et al., 2017] uses attention to replace recurrent dependencies, making the representation at time step $i$ independent from the other time steps. This allows for parallelization of the computation for all time steps in encoder and decoder (see Figure 1b). The general architecture of a decoder accessing a sequence of source encoder states through an attention mechanism remains the same as in Section 2.1. However, the decoder may use multiple encoder-attention mechanisms in each of its layers.

**Positional embeddings** Unlike for RNNs, the independence between time steps requires the explicit encoding of positional information in a sequence. Hence, one-hot vectors $\mathbf{x}_i$ are embedded as $\mathbf{h}_i^0 = \mathbf{E}_S \mathbf{x}_i + \mathbf{e}_{pos,i}$, where $\mathbf{e}_{pos,i}$ is the positional embedding at position $i$. The positional embeddings may be a learned embedding matrix $\mathbf{E}_{pos} \in \mathbb{R}^{e \times n_{max}}$, or be fixed to $\mathbf{e}_{pos,2i} = \sin(pos/10000^{2i/d_{model}})$ and $\mathbf{e}_{pos,2i+1} = \cos(pos/10000^{2i/d_{model}})$ as in Vaswani et al. [2017].

**Encoder** The embedding is followed by several identical encoder blocks consisting of two core sublayers: self-attention and a feed-forward network. The self-attention mechanism is a variation of the dot-product attention [Luong et al., 2015] but generalized to three inputs: a query matrix $\mathbf{Q} \in \mathbb{R}^{n \times d}$, a key matrix $\mathbf{K} \in \mathbb{R}^{n \times d}$, and a value matrix $\mathbf{V} \in \mathbb{R}^{n \times d}$, where $d$ denotes the number of hidden units. Vaswani et al. [2017] further extend attention to multiple *heads*, allowing for focusing on different parts of the input. A single *head* $u$ produces a context matrix

$$\mathbf{C}_u = \text{softmax}\left(\frac{\mathbf{Q}\mathbf{W}_u^Q(\mathbf{K}\mathbf{W}_u^K)^\top}{\sqrt{d_u}}\right)\mathbf{V}\mathbf{W}_u^V, \tag{6}$$

where matrices $\mathbf{W}_u^Q, \mathbf{W}_u^K$ and $\mathbf{W}_u^V$ belong to $\mathbb{R}^{d \times d_u}$. The final context matrix is given by concatenating the heads, followed by a linear transformation: $\mathbf{C} = [\mathbf{C}_1; \ldots; \mathbf{C}_h]\mathbf{W}^O$. The number of hidden units $d$ is chosen to be a multiple of the number of units per head such that $d_u = d/h$. The form in Equation 6 suggests parallel computation across all time steps in a single large matrix multiplication. Given a sequence of hidden states $\mathbf{h}_i$ (or input embeddings), concatenated to $\mathbf{H} \in \mathbb{R}^{n \times d}$, the encoder computes self-attention using $\mathbf{Q} = \mathbf{K} = \mathbf{V} = \mathbf{H}$.

The second subnetwork of an encoder block is a feed-forward network with ReLU activation defined as

$$FFN(\mathbf{x}) = \max(0, \mathbf{x}\mathbf{W}_1 + \mathbf{b_1})\mathbf{W}_2 + \mathbf{b}_2, \tag{7}$$

which again is easily parallelizable across time steps.

Each sublayer, self-attention and feed-forward network, is followed by a post-processing stack of dropout, layer normalization [Ba et al., 2016], and residual connection, such that the output is $LayerNorm(x + Dropout(Sublayer(x)))$. A complete encoder block hence consists of

$$\text{Self-attention} \to \text{Post-process} \to \text{Feed-forward} \to \text{Post-process} \tag{8}$$

and can be stacked to form a multi-layer encoder.

**Decoder** Self-attention in the decoder is applied as in the encoder with $\mathbf{Q} = \mathbf{S}, \mathbf{K} = \mathbf{S}, \mathbf{V} = \mathbf{S}$. To maintain auto-regressiveness of the model, attention on future time steps is masked out accordingly [Vaswani et al., 2017]. In addition to self-attention, a source attention layer which uses the encoder hidden states as key and value inputs is added. Given decoder hidden states $\mathbf{S} \in \mathbb{R}^{m \times s}$ and the encoder hidden states of the final encoder layer $\mathbf{H}^l$, source attention is computed as in Equation 6 with $\mathbf{Q} = \mathbf{S}, \mathbf{K} = \mathbf{H}^l, \mathbf{V} = \mathbf{H}^l$.

The sequence of operations of a single decoder block is:

$$\text{Self-attention} \to \text{Post-process} \to \text{Encoder attention} \to \text{Post-process} \to \text{Feed-forward} \to \text{Post-process}$$

Multiple blocks are stacked to form the full decoder network and the representation of the last block is fed into the output layer of Equation 1.

### 2.3 Fully Convolutional Models (ConvSeq2Seq)

The fully convolutional model [Gehring et al., 2017] also dispenses with recurrent connections and uses convolutions as its core operations. Hence, input embeddings are again augmented with an explicit positional encoding: $\mathbf{h}_i^0 = \mathbf{E}_S \mathbf{x}_i + \mathbf{e}_{pos,i}$.

**Encoder** The convolutional encoder applies a set of (stacked) convolutions that are defined as:

$$\mathbf{h}_i^l = v(\mathbf{W}^l[\mathbf{h}_{i-\lfloor k/2 \rfloor}^{l-1}; \ldots; \mathbf{h}_{i+\lfloor k/2 \rfloor}^{l-1}] + \mathbf{b}^l) + \mathbf{h}_i^{l-1}, \tag{9}$$

where $v$ is a non-linearity and $\mathbf{W}^l \in \mathbb{R}^{d_{cnn} \times kd}$ the convolutional filters. To increase the context window captured by the encoder architecture, multiple layers of convolutions are stacked (also see Figure 1c). The inputs need to be padded with zero vectors at the extremes such that $\mathbf{h}^{l-1}_{-1} = \mathbf{0}, \ldots \mathbf{h}^{l-1}_{\lfloor -k/2 \rfloor} = \mathbf{0}$ and $\mathbf{h}^{l-1}_{n+1} = \mathbf{0}, \ldots \mathbf{h}^{l-1}_{n+\lfloor k/2 \rfloor} = \mathbf{0}$, so that the output consists of as many hidden states as the input. While $v$ can be any non-linearity, Gated Linear Units (GLUs) have been shown to work well in practice [Gehring et al., 2017, Dauphin et al., 2016]. With Gated Linear Units we set $d_{cnn} = 2d$ such that we can split $\mathbf{h} = [\mathbf{h}_A, \mathbf{h}_B] \in \mathbb{R}^{2d}$ and compute the non-linearity as

$$v([\mathbf{h}_A; \mathbf{h}_B]) = \mathbf{h}_A \otimes \sigma(\mathbf{h}_B). \tag{10}$$

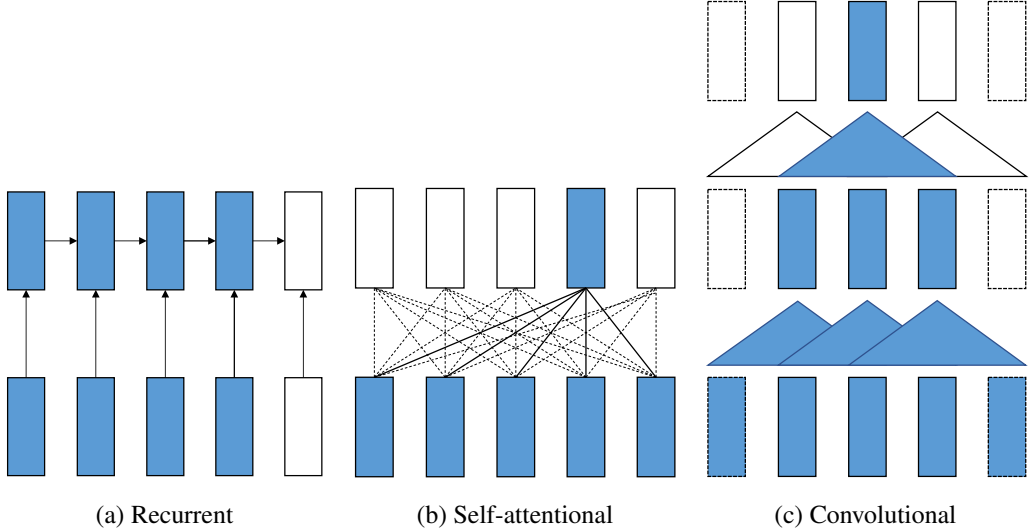

| (a) Recurrent | (b) Self-attentional | (c) Convolutional |

Figure 1: Schematic visualization compute dependencies for the different encoder architectures.

Gated Linear Units can be viewed as a gated variation of ReLUs.

**Decoder** The decoder is similar to the encoder but adds an attention mechanism to every layer. The output of the target side convolution

$$\mathbf{s}^{l*}_t = v(\mathbf{W}^l [\bar{\mathbf{s}}^{l-1}_{t-k+1}; \ldots; \bar{\mathbf{s}}^{l-1}_t] + \mathbf{b}^l) \tag{11}$$

is combined to form $\mathbf{S}^*$ and then fed as an input to the attention mechanism of Equation 6 with a single attention head and $\mathbf{Q} = \mathbf{S}^*, \mathbf{K} = \mathbf{H}^l, \mathbf{V} = \mathbf{H}^l$, resulting in a set of context vectors $\mathbf{c}_t$. The full decoder hidden state is a residual combination with the context such that

$$\bar{\mathbf{s}}^l_t = \mathbf{s}^{l*}_t + \mathbf{c}_t + \bar{\mathbf{s}}^{l-1}_t. \tag{12}$$

To avoid convolving over future time steps at time $t$, the input is padded to the left such that $\mathbf{s}^{l-1}_{-k+1} = \mathbf{0}, \ldots \mathbf{s}^{l-1}_{-1} = \mathbf{0}$.

### 2.4 Context size & Parallelization

As mentioned above, the core difference between the three architectures lies in how the representation $\mathbf{h}^l_i$ of a word $x_i$ in layer $l$ is encoded, given representations from the layer below $\mathbf{h}^{l-1}_1 \ldots \mathbf{h}^{l-1}_n$. This is depicted in Figure 1:

- The recurrent nature of the RNN creates a dependency on the representation of the previous time step, which is the only way to incorporate context from prior representations.

- Self-attention allows a direct information flow from all $\mathbf{h}^{l-1}_1 \ldots \mathbf{h}^{l-1}_n$ to $\mathbf{h}^l_i$ and does not depend on $\mathbf{h}^l_{i-1}$, giving way to fully parallelizing the computation of $\mathbf{h}^l_i$ at the expense of running the attention mechanism over the full sequence at every time step.

- Convolutions also allow for parallelization but limit this window to a fixed size. The effective window size grows as one starts stacking multiple layers. Dashes in Figure 1c denote zero padding vectors, that need to be added in order to produce as many output hidden states as input hidden states.

| Name | score(**s**, **h**) = |
|---|---|
| MLP | $\mathbf{v}_a^\top \tanh(\mathbf{W}_u \mathbf{s} + \mathbf{W}_v \mathbf{h})$ |
| Dot | $\mathbf{s}^\top \mathbf{h}$ |
| Bilinear | $\mathbf{s}^\top \mathbf{W} \mathbf{h}$ |
| Multi-head | $\mathrm{softmax}\left(\frac{\mathbf{s}\mathbf{W}_u^Q (\mathbf{h}\mathbf{W}_u^K)^\top}{\sqrt{d_u}}\right) \mathbf{h}\mathbf{W}_u^V$ |
| Location | $\mathbf{v}_a \mathbf{s}$ |

Table 1: RNN attention types in SOCKEYE.

Note that the parallelization over target sequence time steps is only possible in teacher-forced training, where the full target sequence is known in advance. During decoding or when sampling target sequences, such as in Minimum Risk Training [Shen et al., 2015], each decoder time-step is executed sequentially.

## 3 The SOCKEYE toolkit

In addition to the currently supported architectures introduced in Section 2, SOCKEYE contains a number of model features (Section 3.1), training features (Section 3.2), and inference features (Section 3.3). We will highlight these in the following section, but refer the reader to the public code repository[5] for a more detailed manual on how to use these features and to the references for detailed descriptions.

### 3.1 Model features

**Layer and weight normalization**  To speed up convergence of stochastic gradient descent (SGD) learning methods, SOCKEYE implements two popular techniques, similar in spirit: layer normalization [Ba et al., 2016] and weight normalization [Salimans and Kingma, 2016]. Both apply an affine transformation to the input weights of a neuron. Namely, if a neuron implements a non-linear mapping $f(\mathbf{w}^\top \mathbf{x} + b)$, its input weights $\mathbf{w}$ are transformed, for layer normalization, as $\mathbf{w}_i \leftarrow (\mathbf{w}_i - \mathrm{mean}(\mathbf{w}))/\mathrm{var}(\mathbf{w})$, where the weight mean and variance are calculated over all input weights of the neuron; and, for weight normalization, as $\mathbf{w} = g \cdot \mathbf{v}/||\mathbf{v}||$, where the scalar $g$ and the vector $\mathbf{v}$ are neuron's new parameters.

**Weight tying**  Sharing weights of the input embedding layer and the top-most output layer has been shown to improve language modeling quality [Press and Wolf, 2016] and to reduce memory consumption for NMT. It is implemented in SOCKEYE by setting $\mathbf{W}_o = \mathbf{E}_T$. For jointly-built BPE vocabularies [Sennrich et al., 2017a], SOCKEYE also allows setting $\mathbf{E}_S = \mathbf{E}_T$.

**RNN attention types**  Attention is a core component of NMT systems. Equation 5 gave the basic mechanism for attention for RNN based architectures, but a wider family of functions can be used to compute the score function. Table 1 gives an overview of the attention mechanisms in SOCKEYE.[6]

**RNN context gating**  Tu et al. [2017] introduce context gating for RNN models as a way to better guide the generation process by selectively emphasizing source or target contexts. For example, when generating contents words, the source context should be more relevant, whereas functional target words should be conditioned more on the target context. This is accomplished by introducing a gate

$$\mathbf{z}_t = \sigma(\mathbf{W}_z \mathbf{E}_T(y_{t-1}) + \mathbf{U}_z \mathbf{s}_{i-1} + \mathbf{C}_z \bar{\mathbf{s}}_t), \tag{13}$$

where $\mathbf{W}_z$, $\mathbf{U}_z$ and $\mathbf{C}_z$ are trainable weight matrices. SOCKEYE implements the "both" variant of Tu et al. [2017], where $\mathbf{z}_t$ multiplies the hidden state $\mathbf{s}_t$ of the decoder and $1 - \mathbf{z}_t$ multiplies the source context $\mathbf{h}_t$.

---

[5] https://github.com/awslabs/sockeye/
[6] Note that the transformer and convolutional architectures cannot use these attention types.

**RNN coverage models**   Two well known problems of NMT systems are over- and under-generation. Under-generation occurs when the system leaves part of the input sentence untranslated, while over-generation refers to the problem of generating sequences of target-language-fluent text that is not a translation of any portion of the input. A partial explanation both of these phenomena is that NMT systems do not keep track of which parts of the input sentence have already been translated, in contrast to phrase- and syntax-based systems, which explicitly track the source-language coverage. Tu et al. [2016] mimic this behaviour by keeping track of the attention activations for each source word, and also present some generalizations of this concept, taking advantage of the flexibility of neural approaches. SOCKEYE implements different variants of coverage modeling, ranging from the simpler count models to the more flexible generalizations.

**RNN attention feeding**   When training multi-layer RNN systems, there are several possibilities for selecting the hidden state that is used for computation of the attention score. By default SOCKEYE uses the last decoder RNN layer and combines the computed attention with the hidden RNN state in a feed-forward layer similar to Luong et al. [2015]. Following Wu et al. [2016], SOCKEYE also supports the option to use the first layer of the decoder to compute the attention score, which is then fed to the upper decoder layers.

### 3.2   Training features

**Optimizers**   SOCKEYE can train models using any optimizer from MXNET's library, including stochastic gradient descent (SGD) and the widely-used Adam [Kingma and Ba, 2014]. SOCKEYE also includes its own implementation of the Eve optimizer, which extends Adam by incorporating information from the objective function [Koushik and Hayashi, 2016]. Specifically, Eve applies the following update rule:

$$\theta_t \leftarrow \theta_{t-1} - \alpha \frac{\hat{m}_t}{d_t \sqrt{\hat{v}_t} + \epsilon} \tag{14}$$

As with Adam's update, $\theta$ is the value of model parameters, $\alpha$ is the learning rate, $\hat{m}$ and $\hat{v}$ are bias-corrected moving averages of the gradient and squared gradient, and $\epsilon$ is a very small value. The additional $d$ term is the smoothed moving average of the objective function's relative change. This allows learning to accelerate over flat areas of the loss surface and decelerate when saddle points cause the objective to "bounce".

**Learning schedules**   While optimizers such as Adam and Eve dynamically adjust the step size for each update, recent work has shown the value of annealing the base learning rate $\alpha$ throughout training [Vaswani et al., 2017, Denkowski and Neubig, 2017]. SOCKEYE's 'plateau-reduce' scheduler implements rate annealing as follows. At each training checkpoint, the scheduler compares validation set perplexity against the best previous checkpoint. If perplexity has not surpassed the previous best in $N$ checkpoints, the learning rate $\alpha$ is multiplied by a fixed constant and the counter is reset. Optionally, the scheduler can reset model and optimizer parameters to the best previous point, simulating a perfect prediction of when to anneal the learning rate. For completeness, SOCKEYE also contains a fixed-step scheduler that trains models for a series of fixed intervals with fixed learning rates.

SOCKEYE also includes several stopping criteria. By default, training concludes when validation set perplexity has not improved in $M$ checkpoints, where $M$ is larger than the $N$ required for rate annealing. This gives the optimizer an opportunity to explore with annealed learning rates before concluding that no better point can be found. Training can also be forced to run for a minimum or maximum number of updates or epochs. There is also the choice of optimization metrics on which to predicate early stopping: the user may specify perplexity, accuracy, or BLEU score.

**Monitoring**   Training progress is tracked in a metrics file that contains statistics computed at each checkpoint. It includes the training and validation perplexities, total time elapsed, and optionally a BLEU score on the validation data. To monitor BLEU scores, a subprocess is started at every checkpoint that decodes the validation data and computes BLEU. Note that this is an approximate BLEU score, as source and references are typically tokenized and possibly byte-pair encoded. All statistics can also be written to a Tensorboard event file that can be rendered by a standalone

| Model | Speed (sentences/second) | BLEU |
|---|---:|---:|
| Small | 3.05 | 23.88 |
| Small with vocabulary selection | 5.09 | 23.80 |
| Large | 1.77 | 24.70 |
| Large with vocabulary selection | 2.33 | 24.65 |

Table 2: EN→DE decoding speeds and BLEU score for RNN models with and without vocabulary selection ($K = 100$). Small models use 1 encoder layer and 1 decoder layer while large models use 4 encoder layers and 8 decoder layers.

Tensorboard fork.[7] This allows for easy comparison of learning curves between different experiments and model architectures.

**Regularization** SOCKEYE supports standard techniques for regularization, such as dropout. This includes dropout on input embeddings for both the source and the target and the proposed dropout layers for the transformer architecture. One can also enable *variational dropout* [Gal and Ghahramani, 2016] to sample a fixed dropout mask across recurrent time steps, or *recurrent dropout without memory loss* [Semeniuta et al., 2016]. SOCKEYE can also use MXNET's *label smoothing* [Pereyra et al., 2017] feature to efficiently back-propagate smoothed cross-entropy gradients without explicitly representing a dense, smoothed label distribution.

**Fault tolerance** SOCKEYE saves the training state of all training components after every checkpoint, including elements like the shuffling data iterator and optimizer states. Training can therefore easily be continued from the last checkpoint in the case of aborted process.

**Mult-GPU training** SOCKEYE can take advantage of multipe GPUs using MXNET's data parallelism mechanism. Training batches are divided into equal-sized chunks and distributed to the different GPUs which perform the computations in parallel.[8]

### 3.3 Inference features

SOCKEYE supports beam search on CPUs and GPUs through MXNET. Our beam search implementation is optimized to make use of MXNET's symbolic and imperative API and uses its operators as much as possible to let the MXNET framework efficiently schedule operations. Hypotheses in the beam are length-normalized with a configurable length penalty term as in Wu et al. [2016].

**Ensemble decoding** SOCKEYE supports both linear and log-linear ensemble decoding, which combines word predictions from multiple models. Models can use different architectures, but must use the same target vocabulary.

**Batch decoding** Batch decoding allows decoding multiple sentences at once. This is particularly helpful for large translation jobs such as back-translation [Sennrich et al., 2015], where throughput is more important than latency.

**Vocabulary selection** Each decoding time step requires the translation model to produce a distribution over target vocabulary items. This output layer requires matrix operations dominated by the size of the target vocabulary, $|\mathbf{V}_{trg}|$. One technique for reducing this computational cost involves using only a subset of the target vocabulary, $\mathbf{V}'_{trg}$, for each sentence based on the source [Devlin, 2017]. SOCKEYE can use a probabilistic translation table[9] for dynamic vocabulary selection during decoding. For each input sentence, $\mathbf{V}'_{trg}$ is limited to the top $K$ translations for each source word, reducing the size of output layer matrix operations by 90% or more. As shown in Table 2, this significantly increases decoding speed without quality degradation. The speedup is more pronounced for smaller models where the output layer constitutes more of the total computation.

---
[7] https://github.com/dmlc/tensorboard
[8] It is important to adapt the batch size accordingly.
[9] For example, as produced by `fast_align` [Dyer et al., 2013].

**Attention visualization**   In cases where the attention matrix of RNN models is used in downstream applications, it is often useful to evaluate its soft alignments. For this, SOCKEYE supports returning or visualizing the attention matrix of RNN models.

## 4   Experiments and comparison to other toolkits

We benchmarked each of SOCKEYE's supported architectures against other popular open-source toolkits.[10] These experiments are grouped by architecture. The toolkits we used for each architecture are described briefly here.

*RNN models:*

- OPENNMT [Klein et al., 2017] – a Torch-based NMT toolkit developed at Systran and Harvard University. It includes both Lua and Python implementations and has been used in several production MT applications.
- MARIAN [Junczys-Dowmunt et al., 2016] – a C++ toolkit developed at Adam Mickiewicz University (AMU) in Poznań and at the University of Edinburgh.
- NEMATUS [Sennrich et al., 2017b] – a Theano-based Python toolkit developed at the University of Edinburgh. It has been used to produce some of the best-scoring submissions in the last WMT evaluations.
- NEURALMONKEY [Helcl and Libovický, 2017] – a Tensorflow-based system developed at the Charles University in Prague and the German Research Center for Artificial Intelligence.

*Transformer models:*

- TENSOR2TENSOR (T2T)[11] – a Tensorflow-based sequence-to-sequence toolkit containing the reference implementation for Transformer models [Vaswani et al., 2017].
- MARIAN (as above).

*Convolutional model:*

- FAIRSEQ [Gehring et al., 2017] – the PyTorch-based reference implementation of fully convolutional models.

Cross-system comparisons are difficult. Time and resource constraints prevented us from doing full search across toolkit hyperparameters. Instead, we made a good-faith effort to discover the best settings for each toolkit by looking for recipes included with the toolit, trawling recent papers, or communicating with toolkit authors. This effort represents what a person considering each of the toolkits might do. It is of course likely that the maintainers of each toolkit could find additional areas for improvement. We therefore stress that purpose of these comparisons is to establish coarse-grained characterizations of the different toolkits and architectures, as opposed to definitive statements about which is best.

### 4.1   Data

We ran experiments on two language directions: English into German (EN→DE) and Latvian into English (LV→EN). Models in both language pairs were based on the complete parallel data provided for each task as part of the Second Conference on Machine Translation [Bojar et al., 2017]. Table 3 contains statistics on the data used in our experiments.

The preprocessing scheme was held constant over all models. It was constructed through four steps: normalization, tokenization, sentence-filtering, and byte-pair encoding.

---

[10] All system outputs and training scripts are available at https://github.com/awslabs/sockeye/tree/arxiv_1217.

[11] https://github.com/tensorflow/tensor2tensor

| | EN→DE | | | LV→EN | | |
|---|---|---|---|---|---|---|
| Dataset | Sentences | Tokens | Types | Sentences | Tokens | Types |
| Europarl v7/v8 | 1,905,421 | 91,658,252 | 862,710 | 637,687 | 27,256,803 | 437,914 |
| Common Crawl | 2,394,616 | 97,473,856 | 3,655,645 | - | - | - |
| News Comm. v12 | 270,088 | 11,990,594 | 460,220 | - | - | - |
| Rapid Release 2016 | 1,327,454 | 44,965,622 | 932,855 | 306,381 | 9,357,735 | 318,324 |
| DCEP | - | - | - | 3,528,996 | 64,880,770 | 845,878 |
| Farewell | - | - | - | 9,571 | 145,133 | 29,259 |
| Leta | - | - | - | 15,670 | 779,041 | 107,249 |
| Newstest 2015 | 2,169 | 90,916 | 16,712 | - | - | - |
| Newstest 2016 | 2,999 | 126,828 | 20,845 | - | - | - |
| Newsdev 2017 | - | - | - | 2,003 | 89,545 | 20,646 |
| Newstest 2017 | 3,004 | 125,704 | 20,798 | 2,001 | 86,903 | 19,690 |

Table 3: Data sizes of training (upper block), development (middle block), and test (lower block) sets. All statistics are computed after tokenization and length filtering. Length filtering removed about 0.5% of the data by line count and about 1.3% by word count.

1. *Normalization.* We used Moses' `normalize-punctuation.perl -l LANG` and then removed non-printing characters with `remove-non-printing-char.perl` [Koehn et al., 2007].

2. *Tokenization.* We used Moses' `tokenizer.perl` with the following arguments: `-no-escape -l LANG -protected PATTERNS`, where `PATTERNS` is the basic-protected-patterns file included with the Moses tokenizer. Case was retained as found in the training data. No true-casing was applied.

3. *Filtering* (training only). Sentences longer than 100 tokens on either side were removed using the Moses' `clean-corpus-n.perl` with arguments `1 100`.

4. *Byte-pair encoding.* We trained a byte-pair encoding model with 32,000 split operations [Sennrich et al., 2016].

### 4.2 Evaluation

We calculated our experimental results against the WMT 2017 newstest evaluation set. Our evaluation metric is cased BLEU score [Papineni et al., 2002] with a single reference. We computed all scores by first detokenizing the system output with the Moses detokenization script. The scoring is the same as that used in the WMT evaluation [Bojar et al., 2017]. Consequently, all numbers in this paper can be directly compared against numbers published there.[12]

### 4.3 RNN Experiments

#### 4.3.1 Basic Groundhog model

To begin, we tried our best to build a basic, comparable Groundhog model [Bahdanau et al., 2014] with all chosen frameworks focusing on RNNs. In particular, we configured each framework to train a 1-layer LSTM encoder with bidirectional hidden states concatenated together, MLP attention, 1-layer LSTM decoder and no embedding sharing between encoder and decoder. In addition to the same training and validation data (segmented with the same joint BPE model) and the same word embedding and RNN hidden sizes (500 and 1000, respectively), all frameworks used batch size 80, maximum source and target lengths 50, dropout probability 0.3 in recurrent cells (if supported), gradient norm clipping to 1.0, and the Adam optimizer. Test sets were decoded with a single model using a beam size of 5, source length limit of 100, and a batch size 1. All models were trained on a single Tesla K80 GPU until termination according to the framework's own stopping criterion (when no automatic criterion was available, the best model for testing was picked manually among

---

[12] These are also available online at `matrix.statmt.org`, reported under the *BLEU-cased* column.

| Groundhog model | EN→DE | LV→EN |
|---|---|---|
| OPENNMT-LUA | 19.70 | 10.53 |
| OPENNMT-PY | 18.66 | 9.98 |
| MARIAN | 23.54 | 14.40 |
| NEMATUS | 23.86 | 14.32 |
| NEURALMONKEY | 13.73 | 10.54 |
| SOCKEYE | 23.18 | 14.40 |

Table 4: BLEU scores for Groundhog RNN models on newstest2017.

| Groundhog model | Parameters | Training Speed | | Decoding Speed |
|---|---|---|---|---|
| | | Updates/second | Sentences/second | Sentences/second |
| OPENNMT-LUA | 92.37M | 0.6 | 44 | 4.2 |
| OPENNMT-PY | 87.62M | 1.0 | 82 | 1.6 |
| MARIAN | 93.83M | 1.0 | 80 | 8.0 |
| NEMATUS | 78.54M | 0.4 | 31 | 1.6 |
| NEURALMONKEY | 95.17M | 0.4 | 33 | 0.8 |
| SOCKEYE | 87.83M | 1.1 | 76 | 3.7 |

Table 5: Training speed (updates per second) and throughput (as reported by toolkits, converted to average number of source sentences per second) of EN→DE Groundhog models.

20 epochs). All results were obtained with a single training run of the master branch for respective toolkits' repositories as of mid-October, 2017.

Despite our efforts, the number of parameters reported by the toolkits differed within 10%. See Bojar et al. [2016] for a similar attempt to NMT toolkit comparison and a discussion of its difficulty in practice.

Table 4 shows BLEU scores for EN→DE and LV→EN for each of the Groundhog models. We repeat that these results should be not regarded as representative of a particular tookits' capabilities or their best performance in terms of translation quality. This is only to show what one could get from default codebases with a small but reasonable time investment. Dedicated efforts on any of the toolkits will certainly produce better numbers; however, systematic search for an optimal configuration for each toolkit was beyond the scope of this comparison.

We further analyzed training and decoding speed in Table 5. We observe that SOCKEYE achieves competitive training throughput with the fastest competetitor, MARIAN. However, MARIAN is still ahead in decoding speed.

### 4.3.2 Best Available Settings

In addition to the experiment with the Groundhog model, we ran the RNN frameworks in a "best available" configuration, where we took settings from recent relevant papers that used the toolkit, or communicated directly with toolkit authors. We put in a good-faith effort to discover these settings, but were not able to do extensive experiments over hyperparameters that would characterize efforts by a toolkit's authors. This experiment thus mimicks what an experienced user might be able to do when running the toolkit more or less out of the box for a new language pair.

In all instances, we trained only a single model on the provided parallel data. Techniques shown to further increase accuracy, such as back-translation and model ensembling, were not used.

**OPENNMT-LUA & OPENNMT-PY** We mostly followed the configurations from Levin et al. [2017b,a]: 4-layer LSTM encoder/decoder with attention, hidden and embeddings dimensions set to 1000, input feeding and inter-layer dropout of 0.3 (but without case features and residual connections).

| Toolkit | Layers | EN→DE | LV→EN |
|---|---|---|---|
| OPENNMT-LUA | 4/4 | 22.69 | 13.85 |
| OPENNMT-PY | 4/4 | 21.95 | 13.55 |
| MARIAN | 4/4 | 25.93 | 16.19 |
| NEMATUS | 8/8 | 23.78 | 14.70 |
| NEURALMONKEY | 1/1 | 13.73 | 10.54 |
| SOCKEYE | 4/4 | 25.55 | 15.92 |

Table 6: BLEU scores for RNN models with "best found" settings on newstest2017. The *Layers* column shows the number of encoder and decoder layers. Note that there are variations in complexity of what constitutes a layer in different frameworks.

We trained the systems for 20 epochs with an Adam learning rate of 0.0002 and a batch size of 128. The best model was selected based on reported perplexity scores on the validation set.

**MARIAN** We used the official 1.0.0 release of MARIAN.[13] We took the best settings from the Edinburgh deep RNN system Sennrich et al. [2017a], which builds an 8-layer model with tied embeddings, layer normalization, and skip connections. The hidden layer size was 1024 and the embedding size 512. The models were trained until validation-set perplexity stalled for five consecutive checkpoints.

**NEMATUS** For NEMATUS we also used the network settings described in the WMT 2017 submission Sennrich et al. [2017a]. This comprises 8-layer GRU-based deep architecture[14] with bidirectional connections in each layer, a hidden layer size of 1024, and a word embedding size of 512. Source and target embeddings were tied together.

**NEURALMONKEY** The NEURALMONKEY system follows the system description in [Bojar et al., 2016], which replicated the Groundhog (1-layer LSTM encoder/decoder) model from Section 4.3.1.

**SOCKEYE** We used 4 encoder and 4 decoder layers with LSTM cells and residual connections with a layer size of 1024 hidden units. The embedding size was 512. The attention type was set to MLP. We used a word-based batch size of 4096 and checkpointed the model every 4000 updates. Dropout was added to embeddings with probability 0.1 and to the RNN decoder hidden states with probability 0.2. We used layer normalization, clipped absolute gradient values greater than 1.0, tied the source and target embedding parameters, and used label smoothing with 0.1. The model was optimzied with Adam and an initial learning rate of 0.0002. The learning rate was multiplied by 0.7 whenever perplexity stalled for 8 checkpoints. Training was stopped after 32 checkpoints without improvement in perplexity. At test time, we used a beam size of 5 and a length penalty of 1.0.

The test BLEU scores are given in Table 6. We note once more that the scores should not be taken to indicate peak toolkit performance because of considerable differences in architectures that are well-known to greatly affect BLEU scores (e.g., the number of layers). We show these results solely to indicate rough ballpark performance that one might expect to obtain following recommendation or recipes in published literature without investing in hyperparameter optimization for building a dedicated system.

### 4.4 Self-attentional Transformers

We compared transformer models (Section 2.2) in SOCKEYE and the T2T framework,[15] with respect to BLEU, training throughput, and decoding speed. As we were finalizing these experiments,

---
[13]https://github.com/marian-nmt/marian

[14]For the encoder the 8 layers are divided into 4 forward layers and 4 backward layers. The actual parameter value when using the tool is thus 4.

[15]Version 1.2.7, https://github.com/tensorflow/tensor2tensor/tree/v1.2.7

MARIAN (v1.1.0) also added Transformer models to its feature set, along with a recommended training recipe, as described below.

In order to train on custom user data, the T2T framework requires users to define an implementation of a `Problem` Python class and register it with the codebase. Training and decoding is then carried out by specifying the custom problem name. We tried following the existing EN→DE problem definition for byte-pair encoded WMT'14 data, but obtained significantly worse results than reported in Vaswani et al. [2017]. In particular, output sentences were too short to result in reasonable BLEU scores and we did not manage to find the root cause to this problem. When basing our custom problem definition on the predefined non-BPE problems that build their own sub-word segmentation internally, we obtained good results. For comparability to other experiments and frameworks in this paper, we decided to use the already byte-pair encoded data as input (see Section 4.1), but let T2T build its own subword segmentation with a targeted vocabulary size of 32,000 on top of it. We refer to these models as 'T2T-bpe'. For reference, we also trained a T2T model on tokenized EN→DE data *without* byte-pair encoding to see the effect of impact of working around the issue mentioned above. Such a model, 'T2T-tok', performed only marginally better.

**TENSOR2TENSOR** We used the *transformer_base_v2* hyper-parameter setting[16] which corresponds to a 6-layer transformer with model size of 512. This setting slightly differs from Vaswani et al. [2017] in that it uses dropout by default, fewer learning rate warmup steps (8000), and introduces a pre-process sequence to each layer block (8): each block (self-attention or feed-forward network) is preprocessed with layer normalization, and post-processed with dropout and a residual connection. The total number of trainable parameters was 60,668,928 for EN→DE and 60,484,096 for LV→EN. T2T models were trained on two Volta V100s for a fixed number of 1M global steps using the Adam optimizer, and checkpoints were created every 20 minutes. We did not perform any parameter averaging after training. The batch size was set to approximately 4096 words per batch. The initial learning rate was set to 0.2 and decayed according to the schedule in Vaswani et al. [2017]. At test time, we used a beam size of 5 and a length penalty of 1.0.

**SOCKEYE** We tried to match the T2T settings as close as possible. The SOCKEYE transformer uses the same pre- and post-processing sequence for each sublayer of self-attention and feed-forward network. Models were also trained for 1M updates with a word-based batch size of 4096 using Adam. The learning rate for SOCKEYE was set to 0.0001 and multiplied by a factor of 0.7, whenever validation perplexity did not improve for 8 checkpoints. A checkpoint was saved after every 4,000 updates and no checkpoint averaging was performed after training. The total number of trainable parameters was 62,946,611 for EN→DE and 60,757,127 for LV→EN.

**MARIAN** We used the recommended settings from the Marian transformer tutorial.[17] This also corresponds to a 6-layer transformer with model size of 512 and 8 attention heads. The model also uses tied source and target embeddings and label smoothing. The learning rate was set to 0.0003 with recommended parameters for warm-up and cool-down. Exponential smoothing was used across model checkpoints. The model was trained on four Volta V100s until 10 consecutive checkpoints of non-increasing BLEU score on the validation data. In practice, this was 145,000 updates for the LV→EN model and 365,000 for the EN→DE model. The workspace size was set to 14 MB. At test time, we used a beam size of 5 and a length penalty of 1.0.

**Results** BLEU scores are shown in Table 7. SOCKEYE outperforms the models trained with T2T and is equal to or slightly above MARIAN. It should be noted that these experiments are not aimed at improving state of the art, but illustrate that SOCKEYE achieves competitive BLEU scores without careful hyper-parameter optimization. The difference in the number of parameters between T2T and SOCKEYE for EN→DE comes from fewer embedding parameters in T2T (presumably due to its internal subword) encoding, and the fact that T2T does not seem to use a bias term for the decoder output layer. All transformer parameters for encoder and decoder used the exact same dimensions. We did not perform any other hyper-parameter tuning for the two toolkits.

---

[16]https://github.com/tensorflow/tensor2tensor/blob/v1.2.7/tensor2tensor/models/transformer.py#L664

[17]https://github.com/marian-nmt/marian-examples/tree/ace69bcfd16d30d3a84fd3fd0d879249bf111a85/transformer

| Model | Updates | EN→DE | LV→EN |
|---|---|---|---|
| T2T-bpe | 0.5M | 24.64 | 16.80 |
| T2T-tok | 0.5M | 24.80 | - |
| T2T-bpe | 1M | 26.34 | 17.67 |
| MARIAN | * | 27.41 | 17.58 |
| SOCKEYE | 1M | 27.50 | 18.06 |

Table 7: BLEU scores for Transformer models on newstest2017. MARIAN trained for 415k and 145k updates, respectively.

| Model | Parameters | Training Speed Updates/second | Training Speed Total Training Time | Decoding Speed Sentences/second |
|---|---|---|---|---|
| T2T-bpe | 60.67M | 4.0 | 70 hrs | 15.9 |
| MARIAN | 62.99M | 1.9 | 60 hrs | 13.5 |
| SOCKEYE | 62.95M | 5.2 | 52 hrs | 9.1 |

Table 8: Training speed on 2 Volta V100 on EN→DE data; both models trained for 1M updates/batches. (Marian used 4 Volta V100s and only 415,000 updates). Decoding speed was measured on a single Volta V100 with a batch size of 16 on the WMT'17 EN→DE test set ($n = 3004$). Inputs were sorted by length to batch sentences with similar length.

We also compared training and decoding speeds of the three frameworks (Table 8). SOCKEYE and T2T used a word-based batch size of 4096 during training, while MARIAN used a workspace size of 14 MB. For inference, all three decoders used a sentence-based batch size of 16 on a single Volta GPU. While SOCKEYE trains faster, T2T and MARIAN are faster at inference time.

### 4.5 Fully Convolutional Models

All of our convolutional models used embeddings of size 512 with 8 encoder and decoder layers, which each had 512 hidden units and used convolutional kernels of size 3. The attention mechanism is included on each of the decoder layers.

| Model | Parameters | Words/batch | Training Speed Updates/second | Decoding Speed Sentences/second [18] |
|---|---|---|---|---|
| FAIRSEQ | 81M | ∼3600 | ∼9 | ∼33 |
| SOCKEYE | 86M | ∼3600 | ∼7 | ∼11 |

Table 9: Training speed on 4 Volta V100 on EN→DE data. Decoding speed was measured on 1 Volta V100 with a beam size of 5 and a batch size of 16 on the WMT'17 EN→DE test set ($n = 3004$).

**FAIRSEQ** We used the PyTorch version of FAIRSEQ in version 0.1.0.[19] FAIRSEQ requires a manually compiled PyTorch installation.[20] The training parameters were set equal to the pretrained models available for download.[21] The model contained a total of 81M parameters for EN→DE and 67M for LV→EN. Similar to Gehring et al. [2017] training was done using Nesterov's accelerated gradient method [Sutskever et al., 2013] using a momentum value of 0.99 and clipping gradients norms to a value of 0.1 Dropout with a probability of 0.2 was applied. The mini-batches were allowed to consist of up to 1,000 tokens per GPU, which in practice led to batches of around 3,600 words per batch on 4 GPU. Training was stopped as soon as the validation perplexity did not improve for two epochs. For EN→DE the model converged after 27 epochs and for LV→EN after 24. For inference the checkpoint with the best validation perplexity was used. We decoded with a beam of size 5.

---

[18] We previously reported a decoding speed of 5 sentences/second for fairseq. This measurement was based on its interactive mode, which does not require binarizing the data before decoding but also quietly ignores the batch size flag.

[19] https://github.com/facebookresearch/fairseq-py using commit 30953d8.

[20] Latest commit, 0.2.0+9989bb1, at the time of writing.

[21] https://s3.amazonaws.com/fairseq-py/models/wmt14.en-de.fconv-py.tar.bz2

SOCKEYE  We trained the model with the Adam optimizer with a learning rate of 0.0002, dropout of 0.2, and batches consisting of approximately 4,000 words per batch. A checkpoint was created every 4,000 updates. The model has a total of 86M parameters for EN→DE and 69M for LV→EN. The learning rate was multiplied by a factor of 0.7, whenever the validation perplexity did not improve for 8 checkpoints. Training was stopped as soon as the validation perplexity did not improve for 16 checkpoints. The models converged after 13 epochs for EN→DE and after 11 epochs for LV→EN. For decoding the checkpoint with the best validation perplexity was used. Beam search was performed with a beam size of 5 and a length penalty of 1.0.

| Model   | EN→DE | LV→EN |
| ---     | ---   | ---   |
| FAIRSEQ | 23.37 | 15.38 |
| SOCKEYE | 24.59 | 15.82 |

Table 10: BLEU scores for CNN models on newstest2017.

As shown in Table 10, SOCKEYE models perfom on par with the reference implementation in FAIRSEQ. We additionally carried out training and decoding speed comparisons between the frameworks, which are shown in Table 9. For these, the batch size during training for of SOCKEYE was reduced slightly to match the effective batch size of FAIRSEQ.

### 4.6 Architecture Ensembling

As SOCKEYE implements different architectures within the same framework, ensembling of models can be extended to include different translation models instead of just different parametrizations of the same model as is standard practice. First experiments just using one model for each architecture do not show big improvements over the baseline (see Table 11), but it opens the door for further experimentation.

## 5 Summary

We have presented SOCKEYE, a mature, open-source framework for neural sequence-to-sequence learning that implements the three major architectures for neural machine translation (the only one to do so, to our knowledge). Written in Python, it is easy to install, extend, and deploy; built on top of MXNET, it is fast parallelizable across GPUs. In a set of experiments designed to compare usability and performance of many popular toolkits, we have shown that it achieves competitive performance on WMT tasks (compared to 7 existing NMT toolkits based on 4 different deep learning backends) with minimal investment in terms of setup or hyperparameter optimization. SOCKEYE further implements a broad range of state-of-the-art features, in a codebase conscientiously developed for clarity and extensibility. In the interest of future comparisons and cooperative development through friendly competition, we have provided the system outputs and training and evaluation scripts. We invite feedback and collaboration on the web at https://github.com/awslabs/sockeye.

| System | EN→DE | LV→EN |
|---|---|---|
| SOCKEYE RNN | 24.83 | 15.73 |
| SOCKEYE Convolutional | 24.59 | 15.82 |
| SOCKEYE Transformer | 27.50 | 18.06 |
| SOCKEYE ensemble (RNN, transformer) | 27.72 | 18.80 |

Table 11: BLEU results for single Sockeye systems and ensembles thereof on the WMT'17 newstest evaluation set.